# MSDNN: Multi-Scale Deep Neural Network for Salient Object Detection

Fen Xiao, Wenzheng Deng, Liangchan Peng, Chunhong Cao, Kai Hu, Xieping Gao*

Key Laboratory of Intelligent Computing & Information Processing of Ministry of
Education, Xiangtan University, Hunan 411105, China

*xpgao@xtu.edu.cn

**Abstract.** Salient object detection is a fundamental problem and has been received a great deal of attentions in computer vision. Recently deep learning model became a powerful tool for image feature extraction. In this paper, we propose a multi-scale deep neural network (MSDNN) for salient object detection. The proposed model first extracts global high-level features and context information over the whole source image with recurrent convolutional neural network (RCNN). Then several stacked deconvolutional layers are adopted to get the multi-scale feature representation and obtain a series of saliency maps. Finally, we investigate a fusion convolution module (FCM) to build a final pixel level saliency map. The proposed model is extensively evaluated on four salient object detection benchmark datasets. Results show that our deep model significantly outperforms other 12 state-of-the-art approaches.

## 1 Introduction

Salient object detection, which aims to highlight the most important object regions or distinctive parts in an image, has been a fundamental problem. As a key pre-processing, saliency detection has been applied to benefit various computer vision applications, including image segmentation [1], object detection [2, 3], image retrieval [4], image classification [5], etc. Many computational models [6-12] have been proposed to capture different saliency cues.

Many conventional saliency models rely on various local or global low level features. Local methods [7, 13] aim at evaluating the distinctiveness of each region or pixel with respect to local context for color, texture, and edge orientation, which typically tend to highlight object boundaries while often miss object interiors. Global methods [8, 14] take the entire image into consideration to infer the salient objects. Unlike local methods which are sensitive to high frequency image contents like edges, global methods are often less effective when the textured regions of salient objects are similar to the background. Considering these problems, several researchers integrated local with global contrasts [15, 16] to incorporate their complimentary interactions. However, these methods are weak in capturing high-level semantic information and mainly rely on bottom-up hand-crafted features, which may fail to describe complex image scenarios and object structures, and it is difficult to find the optimal integration.

Convolutional neural networks (CNNs) have powerful capacities in learning complex high-level semantic features of image by establishing deep architectures, which have been successfully used in many applications of computer vision. The CNN-based salient object detection can be divided into region-based models and pixel-based models. Region-based models aim to extract features of each region, and then predict its saliency score. However, exiting region-based methods predict the saliency label of a pixel only considering a limited size of local image patch. They mostly fail to enforce spatial consistency and may inevitably make incorrect predictions. On the other hand, pixel-based models lack the ability to get multi-scope of context for each pixel. An appropriate scope of context is also very important to detect different sizes of objects meanwhile keep those non-salient objects suppressed in the background.

In this paper, we propose a new computational model for saliency detection. The proposed network is capable of extracting hierarchical high-level features and multi-scale contextual simultaneously. First, we use the RCNN to extract global feature representation from the source image. And then, we extend RCNN by introducing upsample layers and exploit FCM to enhance the final saliency map accuracy. Our work has the following contributions:

(1) Multi-scale saliency maps are utilized to improve saliency detection performance. In this framework, we can get different scopes of neighborhood information and robust features, which are effective to detect the small object or multiple objects in the same input image.

(2) FCM has been proposed to fuse the multi-scale saliencies. We use FCM to predict the final pixel level saliency map, which can effectively improve the model performance by integrating different scale feature maps with a nonlinear manner.

## 2 Related works
### 2.1 Convolutional neural network for Salient Object Detection

Inspired by the fact that CNN has forceful ability to extract discriminative features representation from raw pixels automatically, some deep CNN based models [17-20] have been proposed for salient object detection and achieved the state-of-the-art performance. These models can be divided into two categories according to their treatment of input images: region-based methods and pixels-based methods. Region based methods formulate saliency detection as a region classification task. While pixel-based methods directly predict saliency map with end-to-end network. Here, we focus on discussing recent saliency detection models based on deep learning architectures.

For the region-based CNN methods, Zhao et al. [17] developed a multi-context deep learning framework which extracts features of given superpixels at global and local scale, and then predict the saliency score of each region. Considering the fact that the saliency maps are greatly affected by the number of regions, Li et al. [21] presented a CNN approach for salient object detection based on multi-scale feature extraction. These region-based methods lack sufficient context to accurately locate salient objects since they deal with each image regions independently. And the pixel saliency value is given by the mean of pixel saliency values in its enclosing region.

To address the above issues, some works adopt diverse pixel-based deep architectures to preserve details in saliency detection. In [22], Li et al. presented a fully convolutional neural network for saliency detection. Liu et al. [23] proposed a hierarchically refine mechanism from coarse to fine to progressively refine the global saliency map. Although the end-to-end approach enables the model outperforms the region based model in some extent, the spatial information lost in downsampling stage cannot be fully recovered. Recently, some kind of pixel-based information was integrated with higher level region-based information to get the better salient object detection performance. In [19], Yuan et al. proposed a dense and sparse labeling (DSL) framework for saliency detection, both macro object contours and low-level image features were be extracted for saliency estimations. In [20], Tang et al. proposed a novel saliency detection model named as CRPSD, in which the pixel-level CNN and fusion CNN were jointly learned. These models tend to extract local features from each patch and global features from source image. They are very time-consuming, becoming the bottleneck of the computational efficiency of saliency detection.

In this paper, we propose to adopt multi-scale feature representation to enhance the capability of the model to learn discriminative features and boost the details of pixels. The network does not rely on image over-segmentation and post-processing, making the model effective and efficient.

**2.2 Recurrent Convolution Neural Network**

In [24], Liang et al. proposed a RCNN for object recognition, which extract the context information by integrating the recurrent connections into each convolutional layer. As shown in Fig.1, we unfold the recurrent connections for T time steps, the model becomes a deeper feed-forward network with 4(T+1)+2 layers. For a unit located at (i,j) on the kth feature map in an RCL, its net input at time step t is given by:

$$z_{ijk}(t) = (w_k^f)^T u_{(i,j)} + (w_k^r)^T x_{(i,j)}(t-1) + b_k \tag{1}$$

Where $u_{(i,j)}$ and $x_{(i,j)}(t-1)$ denote the feed-forward and recurrent input, respectively, which are the vectorized patches centered at $(i,j)$ of the feature maps in the previous and current layer, $w_k^f$ and $w_k^r$ denote the vectorized feed-forward and recurrent weights, respectively, and $b_k$ is the bias. And its corresponding activity or state is a function of its net input: $x_{ijk}(t) = g(\max(z_{ijk}(t), 0))$.

In Eq.(1), the weight parameters are assumed to be time invariant, we can deepen the network without any extra parameters and keep the computational efficient. And depending on the structure of the RCL, there may be multiple paths between the input layer and the output layer. As a result the receptive filed of an RCL expands when the iteration number increases, which is significant for the network to get more and more context information. In this paper, several stacked RCLs will be used to extract high-level feature representation from the input image.

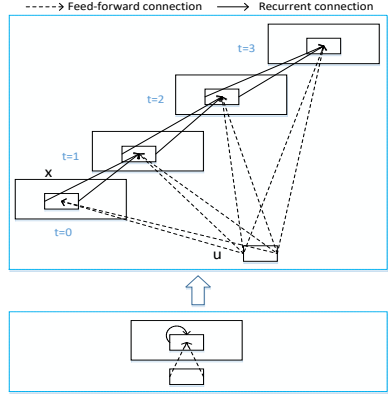

**Fig. 1** The architecture of RCL.

## 3 Proposed Method

The architecture of MSDNN is shown in Fig. 2. First, the RCNN is utilized to extract the high-level context feature fc from global views. Then, hierarchical refined feature maps can be produced based on fc. Next, we get multi-scale top-down saliency maps progressively by exploiting upsampling layers and concatenation layers. At the end of the network, the multi-scale saliency maps are fed into FCM to generate the final saliency map.

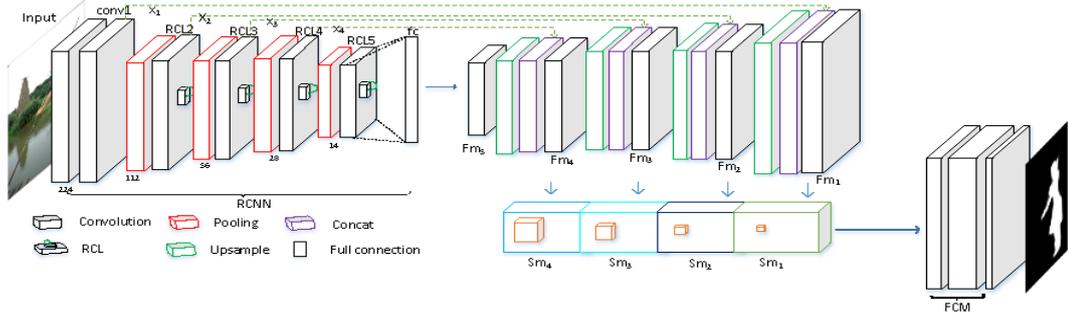

**Fig. 2** The architecture of the proposed neural network.

### 3.1 RCNN for high-level feature extraction

In this subsection, we will use RCNN to extract global features. As shown in Fig 2, the RCNN in our method is made up of six blocks: a convolutional block, four recurrent convolutional blocks and a full connection block. The convolutional block (denoted as conv1 in Fig. 2) consists of two convolution layers and two ReLU [25] layers, in which the number of feature maps is set to 64. Each recurrent block is composed of a pooling layer and a RCL. The size of kernel is defined to 2*2 and the stride is set to 2 in the pooling layer. We set the number of feature maps to 96 and there are 3 recurrent connections in the RCL. A fully connected layer with 6272 nodes is followed by the last recurrent block, which outputs a high-level feature vector fc for the input image. Finally, we reshape and upsample the fc to obtain a coarse global feature map $Fm_5$ (28*28*64), as shown in Fig.3.

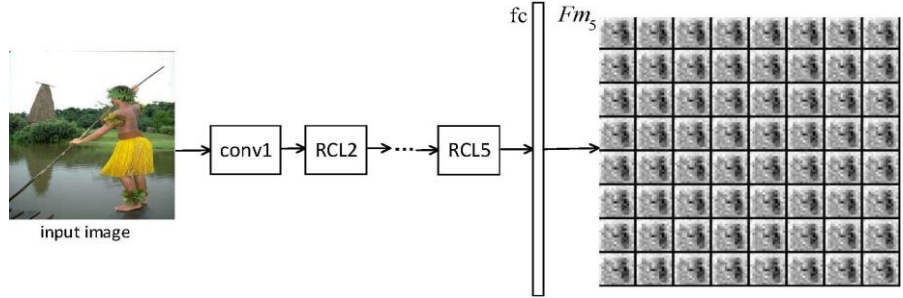

**Fig. 3** Global feature maps obtained by RCNN.

The RCNN is effective in learning global features, while it is not sufficient for saliency detection due to the details maybe lost in down-sample layers. In this paper, multi-scale saliency prior maps will be used to capture local information of salient objects.

### 3.2 Multi-scale saliency maps

In this subsection, feature maps in different scales $Fm_i (i = 4,3,2,1)$ are generated and then integrated with RCL to generate multi-scale saliency prior maps. As shown in Fig. 2, we combine $Fm_5$ with the output of RCL4 (denoted as $X_4$) to generate the refiner feature map $Fm_4$. Similar, we can get the hierarchical feature maps of the input image:

$$Fm_i = Conv(Concat(Fm_{i+1}, X_i); K), i = 4,3,2,1 \qquad (2)$$

Where Concat represents concatenation operator and $X_i$ is the output of corresponding RCL. Conv denotes convolution operator using the kernel K (3*3). Thus, we can get the hierarchical feature maps $Fm_i$, which will be utilized to predict the multi-scale saliency prior maps:

$$Sm_i = Us(Fm_i; S_i), i = 4,3,2,1 \qquad (3)$$

Where Us represents upsample operator. $S_i$ is the upsample kernel with stride of $2^i$, and the sizes of $S_i$ are set to 16*16, 8*8, 4*4 and 1*1, respectively. The upsample operator is exploited to ensure the same spatial size of $Sm_i$ with the original image.

In Fig. 4, we show the detailed framework of multi-scale saliency maps. The hierarchical refined feature maps are exploited to produce multi-scale saliency maps. From $Sm_4$ to $Sm_1$, the receptive fields gradually shrink, and they are representing contextual information at 4 different scales. For each pixel in the saliency map, the network can get multiple scopes of local neighborhood information, which are complementary and can be fused to further improve the performance to detect the small object or multiple objects in the input image. Even for the complex image, the model also can get better effect.

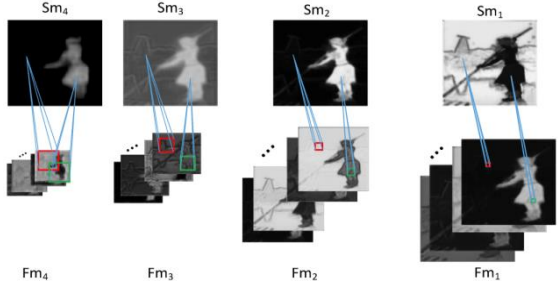

**Fig. 4** Saliency maps with different scales.

**3.3 FCM for multi-scale saliency maps fusion**

Given an input image, the proposed method efficiently produces four saliency prior maps. Finally, FCM exploit the concatenated multi-scale saliency maps to enhance the final salient object accuracy.

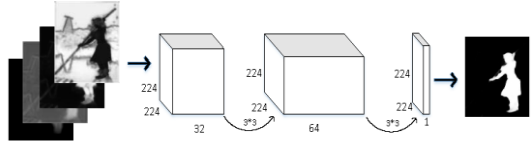

**Fig. 5** The detailed framework of FCM.

The detailed structure of FCM is shown in Fig. 5. The four scale maps are first concatenated into a 4-channel image. Then three convolutional and ReLU layers are used to fuse the feature maps and predict the saliency map. For the former two convolutional layers, we set the number of feature maps to 32 and 64 respectively. Last, we use only one convolutional kernel to convert the features to the final salient map. During training stage, a loss layer is used to compute the errors between the saliency and ground truth with the sigmoid cross-entropy. The FCM and former network are trained jointly to gain the performance improvement and we don't conduct any post-processing.

**4 Experiments**
**4.1 Experimental Setup**
**4.1.1 Datasets**

In this section, we give experimental results to evaluate the performance of MSDNN on four standard benchmark datasets: ECSSD [9], iCoSeg [25], DUT-OMRON [26], MSRA10K [8]. ECSSD is a challenging dataset and includes 1000 semantical meaningful but structurally complex images. iCoSeg was designed for co-segmentation and it contains 643 images and most of the images may contain one or multiple salient objects. DUT-OMRON includes 5168 manually selected high quality images with one or more salient objects and relatively complex backgrounds.

MSRA10K dataset is widely used for saliency detection and covers a large variety of image contents, and most of the images include only salient object with high contrast to the background. All the datasets is bundled with pixel-wise ground truth annotations.

**4.1.2 Evaluation Criteria**

We use standard tools for performance evaluation, precision-recall (P-R) curves, F-measure, mean absolute error (MAE), and area under the ROC curve (AUC), which are most commonly used evaluation metric for saliency detection.

For a saliency map, we generate a series of binary saliency masks by setting the threshold from 0 to 255 with an increment of 5. By comparing the binary masks against the ground truth, pairs of precision and recall scores are obtained. Then we average scores of all images in the database to get the P-R curve. The P-R curve reflects the mean precision and recall of different saliency maps at different thresholds. For quantitative analysis, we binarize each saliency map with an adaptive threshold, which is determined as twice the average value of saliency map. We compute average precision and recall values based on the generated binary maps and the corresponding ground-truth. F-measure score is harmonic mean of precision and recall:

$$F_\beta = \frac{(1+\beta^2)*Precision*Recall}{\beta^2*Precision+Recall} \quad (4)$$

where $\beta^2$ is set to 0.3 as suggested in [9, 27].

And the MAE is the average per-pixel difference between the saliency map S and its corresponding ground truth G:

$$MAE = \frac{1}{W \times H}\sum_{x=1}^{W}\sum_{y=1}^{H}|S(x,y)-G(x,y)| \quad (5)$$

where W and H are width and height of the saliency map, respectively. MAE is more meaningful in evaluating the applicability of a model as it reveals the numerical distance between the saliency map and the ground truth.

Area under the ROC curve (AUC) is one of the most widely used metrics for saliency methods evaluation, which is calculated with Matlab code from Judd's [28] homepage.

**4.1.3 Implementation Details**

We implement the proposed model on the popular open source framework of Caffe [29], it is well known that training such a deep network from scratch is difficult. Therefore, we adapt multi-steps method to train the network in our experiment. In the training state, we resize all the images to 224*224 pixels. The momentums set to 0.9 and the weight decay is 0.0005. In our experiment, we random selected 5500 images from MSRA10K and 2000 images from DUT-OMRON as training set, and selected 500 images from MSRA10K and 500 images from DUT-OMRON as validation, and we test the proposed model on the rest images and the other two datasets. It costs about 11 hours to training the network.

**4.2 Performance Comparison**

We compared the proposed saliency model with 12 state-of-the-art models, including CA [30], DSR [31], GMR [26], GR [32], HS [9], PRGL [15], RBD [33], BL [34], DRFI [35], DHSN [23], MDF [17], DSL [19]. DHSN, MDF and DSL are CNN based methods. For fair comparison, the source codes of these models are obtained from the project site of each algorithm or the benchmark evaluation [27].

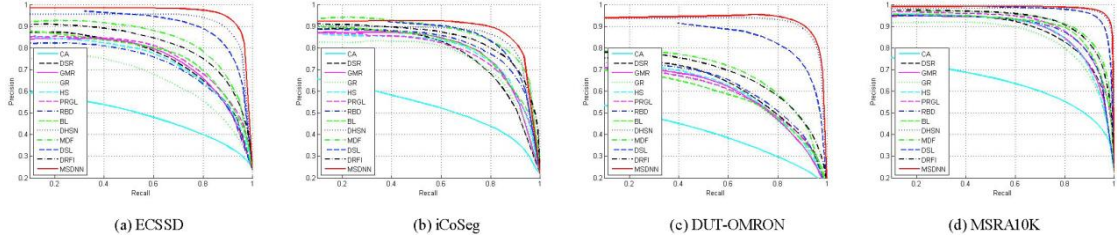

**Fig. 6** Comparison of P-R curves of 12 salient object detection methods on 4 benchmark datasets.

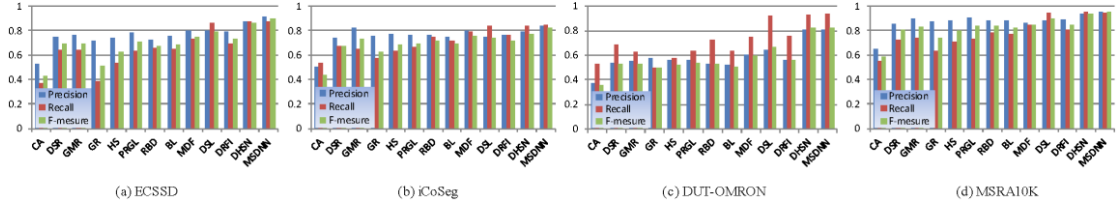

**Fig. 7** Comparison of Precision, Recall and F-measure (using adaptive threshold) among 12different models on 4 datasets.

We compare the P-R curves in Fig. 6. As can be seen that proposed model significantly outperformance the 12 algorithms in terms of the 4 public datasets. The precision achieves maximum value when the recall from 0 to 1 in our model compared with the 12 methods. For the complex and contain multiple objects dataset such as iCoSeg, the proposed method can detect all the objects and obtain a better effect. From the P-R curves we can observe that the precision usually fell down sharply only when the recall is very close to 1. On the contrary, most of other methods usually dropt slowly within the scope of the recall, which shows that the proposed method can generates high contrast saliency maps. Fig. 7 and Fig. 8 shows the precision, recall, F-measure, AUC and MAE of the different models on the 4 datasets. The proposed method shows the highest F-measure and the lowest MAE on all the datasets.

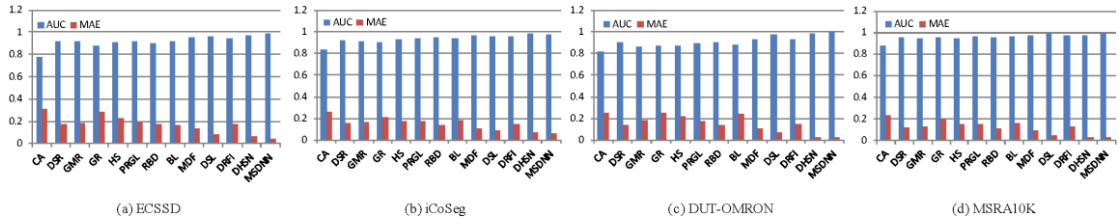

**Fig. 8** Comparison of AUC and MAE (using adaptive threshold) among 12 different models on 4 datasets.

A visual comparison of saliency maps in Fig. 11. As can be seen, the CNN based methods is generally superior to the conventional approaches and the proposed model generates more effect and accurate saliency maps in various challenging cases, e.g., low contrast images, multiple small objects and objects located the image boundary. It is obviously noticed that our saliency maps are very similar to the ground truth and can highlight the saliency regions especially for the simple images such as MSRA10K dataset.

**4.3 Effectiveness of multi-scale saliency maps**

As discussed in Section 3, our multi-scale saliency maps consist of four components, $Sm_4$, $Sm_3$, $Sm_2$ and $Sm_1$. To demonstrate the effectiveness of different scale saliency maps, we trained several models for comparison, $Sm_4$ only, $Sm_{43}$ denote as catenated $Sm_4$ and $Sm_3$, similarity, the proposed denote as the combination of these four components. The qualitative comparison on ECSSD dataset is shown in Fig. 9, the proposed model consistently achieves the best performance, and the model of $Sm_{43}$ is better than $Sm_4$, $Sm_{432}$ is better than $Sm_{43}$. And a visual comparison is given in Fig. 10. The result demonstrates that the proposed multi-scale model not only can get the details information but also can suppress the background noise.

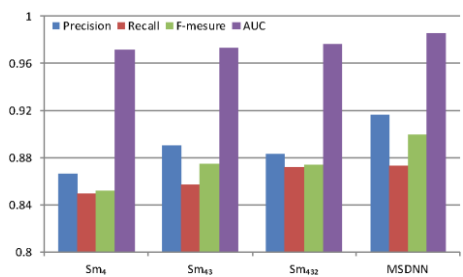

**Fig. 9** Component analysis on ECSSD dataset.

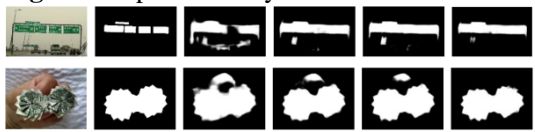

**Fig. 10** Visualization of the multi-scale saliency maps. From left to right: original image, ground truth, $Sm_4$, $Sm_{43}$, $Sm_{432}$, MSDNN.

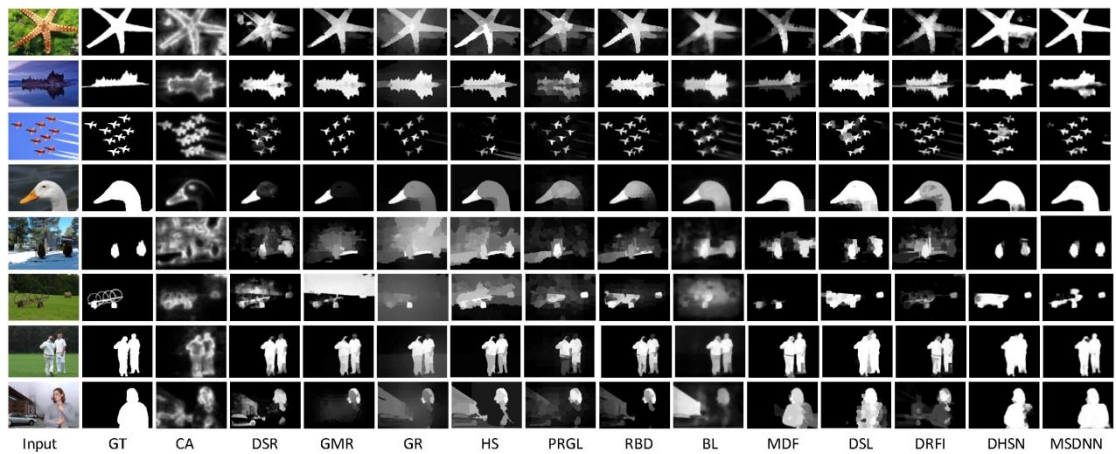

**Fig. 11** Visual comparison of saliency maps from state-of-the-art methods. From top to bottom: ECSSD (first and second rows), iCoSeg (third and fourth rows), DUT-OMRON (fifth and sixth rows), MSRA10K (last two rows). From left to right: input image, ground truth, CA, DSR, GMR, GR, HS, PRGL, RBD, BL, MDF, DSL, DRFI, DHSN, MSDNN.

## 5 Conclusions

In this paper, we proposed a novel method of multi-scale deep supervised neural network for salient object detection. The model exploits RCNN to learn discriminative information and multi-scale image feature representation are adopted to hierarchically reconstruct final saliency map. Experimental demonstrate that our deep model is effective and efficient and can significantly improve the state-of-the-art.